\documentclass[letterpaper, 10 pt, conference]{ieeeconf}
\IEEEoverridecommandlockouts                              

\usepackage{ifpdf}

\usepackage{adjustbox}
\usepackage{xcolor}
\usepackage{color}
\usepackage{bbding}
\usepackage{textcomp}
\usepackage{makecell,rotating,multirow,diagbox}
\usepackage{booktabs,caption}
\usepackage[flushleft]{threeparttable}
\usepackage[font=small]{subcaption}
\usepackage{amsmath}
\usepackage{esvect}
\usepackage{mathtools}
\usepackage{amssymb}
\usepackage{amsfonts}

\usepackage{array}
\usepackage{algorithm}
\usepackage{algorithmic}
\usepackage{caption}
\usepackage{booktabs}
\usepackage{listings}
\usepackage{cite}

\definecolor{lightgray}{gray}{0.9}
\begin{document}
\captionsetup[figure]{labelfont={bf},labelformat={default},labelsep=period,name={Fig.}}
\title{\LARGE \bf
Inequality Constrained Trajectory Optimization with A Hybrid Multiple-shooting iLQR
}

\author{$\text{Yunxi Tang}^{\text{a}
}$, $\text{Xiangyu Chu}^{\text{a}
}$, $\text{Wanxin Jin}^{\text{b}
}$, and $\text{K. W. Samuel Au}^{\text{a}}$ 
\thanks{$^{\text{a}}\text{Department}$ of Mechanical and Automation Engineering, The Chinese University of Hong Kong, Hong Kong, China.
}
\thanks{$^{\text{b}}\text{Department}$ of Mechanical Engineering and Applied Mechanics, University of Pennsylvania, PA 19104, USA.
}
}
\maketitle
\thispagestyle{empty}
\pagestyle{empty}

\begin{abstract}
Trajectory optimization has been used extensively in robotic systems. In particular, iterative Linear Quadratic Regulator (iLQR) has performed well as an off-line planner and online nonlinear model predictive control solver, with a lower computational cost. However, standard iLQR cannot handle any constraints or perform reasonable initialization of a state trajectory. In this paper, we propose a hybrid constrained iLQR variant with a multiple-shooting framework to incorporate general inequality constraints and infeasible states initialization. The main technical contributions are twofold: 1) In addition to inheriting the simplicity of the initialization in multiple-shooting settings, a two-stage framework is developed to deal with state and/or control constraints robustly without loss of the linear feedback term of iLQR. Such a hybrid strategy offers fast convergence of constraint satisfaction. 2) An improved globalization strategy is proposed to exploit the coupled effects between line-searching and regularization, which is able to enhance the numerical robustness of the constrained iLQR approaches. Our approach is tested on various constrained trajectory optimization problems and outperforms the commonly-used collocation and shooting methods.
\end{abstract}

\section{Introduction}
Trajectory optimization has been well studied in control and robotic communities. By minimizing objective functions and incorporating constraints, an implicit and high-level formulation exploits the underlying dynamical properties and achieve eﬀicient behaviors. 
Mayne \cite{Mayne1966} proposed \textit{Differential Dynamic Programming} (DDP) to solve unconstrained TO problems in 1960s. DDP and its variant iterative Linear Quadratic Regulator (iLQR) \cite{iLQR2004} are similarly based on Bellman’s Principle of Optimality \cite{Kelly2017}. Compared with DDP method, iLQR ignores the second-order derivatives of system dynamics, which reduces the computational complexity significantly. iLQR effectively solves TO through parameterized control variables, which requires an initial control policy to generate a stable state trajectory via forward integration. Since states are not optimized directly \cite{Lantoine2012} in classical iLQR, it can hardly be warm-started with dynamically infeasible state trajectories. To overcome these initialization issues, \cite{GiftthalerIROS2018} proposed an unconstrained multiple-shooting (MS) framework for the iLQR approach. Similarly, \cite{MastalliICRA2020} proposed a feasibility-prone DDP (FDDP) variant with taking into account of contacts in legged locomotion. Within the FDDP framework, \cite{MartiIROS2020} considered box constraints on control inputs. The multiple shooting variants show a better flexibility in initialization and distribute nonlinearity by introducing extra state decision variables. Although these works enjoyed the advantages of MS settings and iLQR-like methods, they can only handle specific constraints, such as box-input constraints or holonomic contact constraints. 
\begin{table}[b]
    \caption{Literature Review and Comparison}
	\label{table:review}
	\centering
	\begin{adjustbox}{max width=85mm}
	\begin{tabular}{c|cccc}
		\toprule  
		\textbf{Algorithm} &\textbf{F. P.$^{\rm 1}$}  &\textbf{S. C.$^{\rm 2}$} &\textbf{C. C.$^{\rm 3}$} &\textbf{I. I.$^{\rm 4}$}\\ 
		\midrule 
		DDP\cite{Mayne1966}, iLQR\cite{iLQR2004} & \Checkmark  &\XSolidBrush &\XSolidBrush & \XSolidBrush\\
		\midrule
		iLQR-GNMS\cite{GiftthalerIROS2018}, FDDP\cite{MastalliICRA2020}   &\Checkmark  &\XSolidBrush    & \XSolidBrush &\Checkmark \\
		\midrule 
		sb-FDDP\cite{MartiIROS2020} &\Checkmark &\XSolidBrush &\Checkmark &\Checkmark\\
		\midrule
		CLDDP\cite{TassaICRA2014} &\Checkmark &\XSolidBrush &\Checkmark &\XSolidBrush \\
		\midrule
		CDDP\cite{XieICRA2017} &\XSolidBrush &\Checkmark &\XSolidBrush &\XSolidBrush \\
		\midrule
         (AL)-S-KKT\cite{Yuichiro2020}, IPDDP\cite{PavlovTCST2021} &\Checkmark &\Checkmark &\Checkmark &\XSolidBrush \\
        \midrule
        BCL-DDP\cite{Sarah2021} &\Checkmark &\Checkmark &\XSolidBrush &\Checkmark\\
        \midrule
        ALTRO\cite{HowellIROS2019} &\XSolidBrush &\Checkmark &\Checkmark &\Checkmark\\
		\midrule
		Proposed HM-iLQR &\Checkmark &\Checkmark &\Checkmark &\Checkmark\\
		\bottomrule 
	\end{tabular}
	\end{adjustbox}
    \begin{tablenotes}
      \item $^{\rm 1} \text{feedback policy}$ $^{\rm 2} \text{state constraint}$ $^{\rm 3} \text{control constraint}$ $^{\rm 4} \text{dynamical infeasible initialization}$
    \end{tablenotes}
\end{table}
The effort of extending iLQR to generic constrained problems also has been made. 
Unlike only focusing on either control constraints \cite{TassaICRA2014} or state constraints \cite{XieICRA2017}, \cite{Yuichiro2020} \cite{PavlovTCST2021} dealt with both state and control constraints with Karush–Kuhn–Tucker (KKT) conditions, which solved a series of quadratic problems (QP) in the forward pass to guarantee feasibility. \cite{Sarah2021} \cite{LantoineAIAA2008} combined DDP with Augmented Lagrangian approaches (AL-DDP). \cite{HowellIROS2019}  proposed a hybrid method, ALTRO, which employed AL approaches firstly and post-processed the obtained solution with a direct active-set projection method. Similarly, \cite{Yuichiro2020} applied AL and polished the coarse solution with an interior-point method with slack variables (s-KKT). However, methods in \cite{Yuichiro2020,Sarah2021,LantoineAIAA2008} either get numerical instability easily, or increase the complexity by introducing factorization and projections \cite{TassaICRA2014}\cite{Yuichiro2020}\cite{HowellIROS2019}, or sacrifice the linear feedback policy in the optimal solution \cite{XieICRA2017}\cite{HowellIROS2019}. 
The detailed comparisons between different algorithms are listed in Table \ref{table:review}.
Hence, a generic algorithm framework, which can maintain the advantages of Multiple-shooting iLQR (M-iLQR) and constraint-handling, is desirable.

In this paper, we propose a hybrid multiple-shooting iLQR (HM-iLQR) variant to solve state and/or control constrained trajectory optimization problems. The main contributions are twofold. First, leveraging the simplicity of the initialization in MS settings, we propose a two-stage framework to handle general inequality constraints without loss of the linear feedback term. The combination of penalty method and interior-point method makes the hybrid algorithm achieve robust and fast constraint satisfaction. AL approach has fast convergence at the first several iterations and obtains a coarse solution rapidly. Then an extension of interior-point method, Relaxed Logarithmic Barrier (RLB) approach, is used to eliminate the remained constraint violations of the coarse solution, which also avoids the ill-conditioned problems \cite{Berts2014} in the AL stage. Both stages share similar computational implementations and do not introduce any factorization or projection.
Second, we propose an novel globalization strategy which exploits the coupled effects of line-searching and regularization. Although regularization or line-searching have been used in previous works, few has investigated the lagged effect of regularization on the next forward pass. A large regularization would result higher feedback policy in the next forward pass, which makes the state trajectory update more conservative. Hence, the regularization in the backward pass can be also utilized to control the update step of states, especially when the cost is close to the minimum. The proposed globalization strategy can improve refinements of the coarse solution in the RLB stage by avoiding too small line-search factors.

The paper is organized as follows: Section II recalls the preliminaries of the multiple-shooting iLQR framework. Section III presents the proposed hybrid algorithm in detail. In Section IV, we numerically validate the proposed approach on several constrained TO problems and compare the performances with other TO methods.

\section{Background} 
\subsection{iLQR Preliminaries}
Consider a discrete finite-time optimal control problem,
\begin{equation} \label{1}
    \begin{aligned}
     \min \limits_{\boldsymbol{U}}  \; J(\boldsymbol{x}_{0},\boldsymbol{U}) &= \sum_{k=0}^{N-1} \ell (\boldsymbol{x}_k,\boldsymbol{u}_k) + \ell _{f}(\boldsymbol{x}_N) \\
     s.t.  \;\; \boldsymbol{x}_{k+1} &= \boldsymbol{f}(\boldsymbol{x}_k,\boldsymbol{u}_k), \; k=0,1,2,...,N-1
\end{aligned}
\end{equation}
where $\boldsymbol{x}_k \in \mathbb{R}^n$ and $\boldsymbol{u}_k \in \mathbb{R}^m$ denote the state and control at time $t_k$, respectively. $\boldsymbol{f}: \mathbb{R}^n \times \mathbb{R}^m \mapsto \mathbb{R}^n$ is the transition dynamics. $\boldsymbol{x}_{0}$ is the initial state. The scalar-valued functions $\ell$ and $\ell_f$ denote the running and terminal objective functions. Let $\boldsymbol{X}=(\boldsymbol{x}_0,\dots,\boldsymbol{x}_N)$, $\boldsymbol{U}=(\boldsymbol{u}_0,\dots,\boldsymbol{u}_{N-1})$ be the state and control sequences over the horizon $N$. The value function $V_k$ is the optimal cost-to-go function starting at $\boldsymbol{x}_k$,
\begin{equation} \label{2}
    V_k = \min \limits_{\boldsymbol{u}_k,\dots,\boldsymbol{u}_{N-1}} \; \sum_{i=k}^{N-1} \ell (\boldsymbol{x}_i,\boldsymbol{u}_i) + \ell _{f}(\boldsymbol{x}_N).
\end{equation}
Based on Bellman's Principle of Optimality, the optimal value function can be evaluated recursively,
\begin{equation} \label{3}
     V_k = \min \limits_{\boldsymbol{u}_k} \ell (\boldsymbol{x}_k,\boldsymbol{u}_k) + V_{k+1}\left(\boldsymbol{f}(\boldsymbol{x}_k,\boldsymbol{u}_k)\right).
\end{equation}
The terminal state value $V_N = \ell _f(\boldsymbol{x}_N)$. iLQR performs backward pass and forward pass along nominal state/control trajectory iteratively.
\subsubsection{Backward Pass}
$Q:\mathbb{R}^n \times \mathbb{R}^m  \mapsto \mathbb{R}$ function evaluates the \textit{cost-to-go} of taking an action at a given state,
\begin{equation}
    Q_k(\boldsymbol{x},\boldsymbol{u}) = l(\boldsymbol{x},\boldsymbol{u})+V_{k+1}\left(\boldsymbol{f}(\boldsymbol{x},\boldsymbol{u})\right).
\end{equation}
With the second-order approximation of $Q$, the optimal control modification $\delta {\boldsymbol{u}}^*$ is obtained by solving the following quadratic problem \textit{w.r.t.}, $\delta{\boldsymbol{u}}$:
\begin{equation*}\label{qp}
     \mathop{\arg\min_{\delta \boldsymbol{u}}} \; \frac{1}{2} \left[ \begin{array}{c}
         1\\
         \delta \boldsymbol{x}\\
         \delta \boldsymbol{u}
    \end{array} \right]^T
    \left[
    \begin{array}{ccc}
        0                   & Q^T_{\boldsymbol{x}}    & Q^T_{\boldsymbol{u}}\\
        Q_{\boldsymbol{x}}  & Q_{\boldsymbol{xx}}     & Q_{\boldsymbol{xu}}\\
        Q_{\boldsymbol{u}}  & Q_{\boldsymbol{ux}}     & Q_{\boldsymbol{uu}}
    \end{array}
    \right]
    \left[ \begin{array}{c}
         1\\
         \delta \boldsymbol{x}\\
         \delta \boldsymbol{u}
    \end{array} \right].
\end{equation*}
where,
\begin{subequations}\label{BP}
\begin{gather}
     Q_{\boldsymbol{x},k}  = \ell_{\boldsymbol{x},k}+\boldsymbol{f}_{\boldsymbol{x},k}^T V_{\boldsymbol{x},k+1}, \\
     Q_{\boldsymbol{u},k} = \ell_{\boldsymbol{u},k}+\boldsymbol{f}_{\boldsymbol{u},k}^T V_{\boldsymbol{x},k+1}, \\
     Q_{\boldsymbol{xx},k} = \ell_{\boldsymbol{xx},k}+\boldsymbol{f}_{\boldsymbol{x},k}^T V_{\boldsymbol{xx},k+1}\boldsymbol{f}_{\boldsymbol{x},k}, \\
     Q_{\boldsymbol{uu},k} = \ell_{\boldsymbol{uu},k}+\boldsymbol{f}_{\boldsymbol{u},k}^T V_{\boldsymbol{xx},k+1}\boldsymbol{f}_{\boldsymbol{u},k}, \\
     Q_{\boldsymbol{ux},k} = \ell_{\boldsymbol{ux},k}+\boldsymbol{f}_{\boldsymbol{u},k}^T V_{\boldsymbol{xx},k+1}\boldsymbol{f}_{\boldsymbol{x},k}, \\
     Q_{\boldsymbol{xu},k} = \ell_{\boldsymbol{xu},k}+\boldsymbol{f}_{\boldsymbol{x},k}^T V_{\boldsymbol{xx},k+1}\boldsymbol{f}_{\boldsymbol{u},k}.
\end{gather}
\end{subequations}
$\boldsymbol{f}_{\boldsymbol{x}}$ and $\boldsymbol{f}_{\boldsymbol{u}}$ are the sensitives \textit{w.r.t} the state and control. A local linear control update is obtained by solving (\ref{qp}),
\begin{subequations}\label{control_update}
\begin{gather}
    \delta{\boldsymbol{u}}^*=\boldsymbol{k}_{ff}+\boldsymbol{K}_{fb}\delta{\boldsymbol{x}}, \\
    \boldsymbol{k}_{ff}=-Q_{\boldsymbol{uu}}^{-1}Q_{\boldsymbol{u}}, \;
    \boldsymbol{K}_{fb}=-Q_{\boldsymbol{uu}}^{-1}Q_{\boldsymbol{ux}}.
\end{gather}
\end{subequations}
Plugging $\delta{\boldsymbol{u}}^*$ back into $Q$, back propagation of the derivatives of the value function are as follows:
\begin{subequations}
\begin{align}
    \delta V_{k} & = -\frac{1}{2}Q_{\boldsymbol{u},k}Q^{-1}_{\boldsymbol{uu},k}Q_{\boldsymbol{u},k}, \\
    V_{\boldsymbol{x},k} &= Q_{\boldsymbol{x},k} - Q_{\boldsymbol{u},k}Q^{-1}_{\boldsymbol{uu},k}Q_{\boldsymbol{ux},k}, \\
    V_{\boldsymbol{xx},k} & = Q_{\boldsymbol{xx},k}  - Q_{\boldsymbol{xu},k}Q^{-1}_{\boldsymbol{uu},k}Q_{\boldsymbol{ux},k}.
\end{align}
\end{subequations}
\subsubsection{Forward Pass}
Integrating the dynamics with the updated feedforward and feedback control policy in (\ref{control_update}) yields,
\begin{equation*}
    \begin{aligned}
        \hat{\boldsymbol{u}}_k = \boldsymbol{u}_k + &\boldsymbol{k}_{ff,k} + \boldsymbol{K}_{fb,k}(\hat{\boldsymbol{x}}_k - \boldsymbol{x}_k),\\
        \hat{\boldsymbol{x}}_{k+1} &= \boldsymbol{f}(\hat{\boldsymbol{x}}_k, \;\hat{\boldsymbol{u}}_k),\; \hat{\boldsymbol{x}}_0 = \boldsymbol{x}_0,
    \end{aligned}
\end{equation*}
where $(\hat{\boldsymbol{X}},\hat{\boldsymbol{U}})$ is the improved state/control nominal sequences.
\subsection{Generalization to Multiple Shooting Framework}
The multiple-shooting variant was proposed by introducing intermediate state decision variables \cite{Lantoine2012}. Additional matching constraints \cite{GiftthalerIROS2018} or additional cost term \cite{HowellIROS2019} are required to ensure the dynamical feasibility. Here we give a brief description of the former strategy in \cite{GiftthalerIROS2018} and readers can refer to \cite{GiftthalerIROS2018}\cite{MastalliICRA2020} for more details.
\begin{algorithm}[htb]
    \caption{\lstinline{M-iLQR} in \cite{GiftthalerIROS2018}}
    \label{alg:mddp}
    \begin{algorithmic}[1]
        \REQUIRE \small{$\boldsymbol{X}^{init}_{node}$, $\boldsymbol{U}_{init}$, $N$, $M$, $MaxIter$, $\epsilon_v$, $d_{max}$}\\
        (\small{$MaxIter$: maximum iteration; $\epsilon_v$: threshold of objective value reduction; $d_{max}$: tolerance of state defects})\\
        \STATE \small{\textbf{Initialize}}: $\hat{J} \gets 0$, $J \gets 0$, $L \gets (N-1)/M+1$
        \STATE \small{\textbf{Initial Forward Pass}}:
        \FOR{$i=1 \to M$}
            \STATE $\boldsymbol{X}^{i}[0] \gets \boldsymbol{X}^{init,i}_{node}[0]$
            \FOR{$j=1 \to L-1$}
            \STATE $\boldsymbol{X}^i[j+1] \gets \boldsymbol{f}(\boldsymbol{X}^i[j], \boldsymbol{U}^i_{init}[j])$
            \ENDFOR
        \ENDFOR
        \STATE Compute defects $\boldsymbol{d}$ and initial cost $J_0$, set $J \gets J_0$
        \STATE \textbf{Main\; Loop}:
        \FOR{$iter=1 \to MaxIter$}
            \STATE \small{\textbf{Backward Pass}}:
            \STATE $[\boldsymbol{k}_{ff},\boldsymbol{K}_{fb},\Delta V] \gets$ \lstinline{Backward_Pass} $(\boldsymbol{X}, \boldsymbol{U})$
            \STATE \small{\textbf{Forward Pass}}: 
            \STATE Update $\boldsymbol{X}_{node}$
            \STATE $[\hat{\boldsymbol{X}},\hat{\boldsymbol{U}},\hat{J}] \gets $ \lstinline{Backward_Pass} $ (\boldsymbol{k}_{ff},\boldsymbol{K}_{fb},\boldsymbol{X},\boldsymbol{U},\Delta V,J)$
            \STATE Compute defects $\hat{\boldsymbol{d}}$ and $\Delta J = J -\hat{J}$ 
            \IF{ $\Delta J <\epsilon_v$ and $ |\hat{\boldsymbol{d}}|_2 < d_{max}$}
                \STATE \textbf{return} $\boldsymbol{X}_{sol} \gets \hat{\boldsymbol{U}},\;\boldsymbol{X}_{sol} \gets \hat{\boldsymbol{U}},\;\boldsymbol{K}_{sol} \gets \boldsymbol{K}_{fb}$
            \ENDIF
            \STATE $J \gets \hat{J}$, $\boldsymbol{X} \gets \hat{\boldsymbol{X}}$, $\boldsymbol{U} \gets \hat{\boldsymbol{U}}$
        \ENDFOR
        \RETURN $\boldsymbol{X}_{sol}$, $\boldsymbol{U}_{sol}$ and $\boldsymbol{K}_{sol}$
    \end{algorithmic}
\end{algorithm}
\subsubsection{State Defect}
By breaking a long trajectory into $M$ segments, with a length of $L$ for each phase, the starting state of each segment is a \textit{node} state \cite{GiftthalerIROS2018}. The \textit{defect} between the end state of previous sub-trajectory and the node of next successive segment, $\boldsymbol{x}_{k+1}$, is defined as,
\begin{equation*}
    \boldsymbol{d}_{k} =\boldsymbol{f}(\boldsymbol{x}_k,\boldsymbol{u}_k)-\boldsymbol{x}_{k+1}.
\end{equation*}
The linearized dynamical constraint at the node state is,
\begin{equation}
    \delta \boldsymbol{x}_{k+1} = \boldsymbol{f}_{\boldsymbol{x},k} \delta \boldsymbol{x}_k + \boldsymbol{f}_{\boldsymbol{u},k} \delta \boldsymbol{u}_k + \boldsymbol{d}_k.
\end{equation}
Taking into account of the defects, the modified TO is,
\begin{equation} \label{4}
    \begin{aligned}
    & \min \limits_{\boldsymbol{U}, \boldsymbol{X}_{node}}  \; J(\boldsymbol{X}_{node},\boldsymbol{U}) = \sum_{k=0}^{N-1} \ell (\boldsymbol{x}_k,\boldsymbol{u}_k) + \ell _{f}(\boldsymbol{x}_N) \\
    & s.t.  \;\; \delta \boldsymbol{x}_{k+1} = \boldsymbol{f}_{\boldsymbol{x},k} \delta \boldsymbol{x}_k + \boldsymbol{f}_{\boldsymbol{u},k} \delta \boldsymbol{u}_k + \boldsymbol{d}_k, \\
    & \;\;\;\;\;\;\; \boldsymbol{X}_{node} = \boldsymbol{X}^{init}_{node},
\end{aligned}
\end{equation}
where $\boldsymbol{d}_k$=$\boldsymbol{0}$ if $\boldsymbol{x}_{k+1}$ is not a node. $\boldsymbol{X}_{node}$ is the state decision variables, which is initialized as $\boldsymbol{X}^{init}_{node}$. 
The modified recursion in the backward pass is,
\begin{subequations}{\label{mdp}}
\begin{align}
    Q_{\boldsymbol{x},k} & = \ell_{\boldsymbol{x},k}+\boldsymbol{f}_{\boldsymbol{x},k}^T V^+_{\boldsymbol{x},k+1}, \\
    Q_{\boldsymbol{u},k} & = \ell_{\boldsymbol{u},k}+\boldsymbol{f}_{\boldsymbol{u},k}^T V^+_{\boldsymbol{x},k+1}, 
\end{align}
\end{subequations}
where 
$V^+_{\boldsymbol{x},k+1} = V_{\boldsymbol{x},k+1} + V_{\boldsymbol{xx},k+1}\boldsymbol{d}_k$
is the Jacobian of the state-value function.
The Hessian matrix $V_{\boldsymbol{xx}}$ keeps the same as in (\ref{BP}c-\ref{BP}f).

\subsubsection{State and Control Update}
The control update keeps the same as (\ref{control_update}). 
Based on the linearization of system dynamics, the node states are updated as below,
\begin{subequations}{\label{mfi}}
\begin{gather}
    \hat{\hat{\boldsymbol{x}}}_{k+1}  = \boldsymbol{x}_{k+1} + \delta{\hat{\boldsymbol{x}}_{k+1}}, \\
    \delta{\hat{\boldsymbol{x}}_{k+1}} = \boldsymbol{f}_{\boldsymbol{x},k}\delta \boldsymbol{x}_k+\boldsymbol{f}_{\boldsymbol{u},k}\delta \boldsymbol{u}^*+\boldsymbol{d}_k.
\end{gather}
\end{subequations}
Starting from node $\hat{\boldsymbol{x}}_{k+1}$, we can rewrite the sub-trajectory of each shooting segment via forward integration. The detailed algorithm of M-iLQR is outlined in Algorithm \ref{alg:mddp}.
\section{HM-iLQR: Hybrid Multiple-shooting Iterative Linear Quadratic Regulator} 
In this section, we present our main contribution: a hybrid multiple-shooting iterative Linear Quadratic Regulator (HM-iLQR) approach, to deal with both state and control inequality constraints and retain the linear feedback policy. This hybrid algorithm is composed of two-phase strategies. Firstly, we utilize M-iLQR with the Augmented Lagrangian method to obtain a coarse solution rapidly. Then, the coarse solution is used to warm-start the second stage which uses a relaxed log barrier function. The second stage eliminates the constraint violations. Since constraints are penalized as additional objective terms in both stages, the linear feedback policy can be naturally reserved. We consider additional inequality constraints on the basis of problem in (\ref{2}). 
\begin{equation} \label{yy}
    \boldsymbol{g}(\boldsymbol{x}_k + \delta \boldsymbol{x}_k, \boldsymbol{u}_k + \delta \boldsymbol{u}_k)  \leq \boldsymbol{0},
\end{equation}
where $\boldsymbol{g} \in \mathbb{R}^{q_k}$ is the inequality constraints, $q_k$ is the number of constraints, and $(\boldsymbol{x}_k, \boldsymbol{u}_k)$ is the nominal trajectory. We will show how to incorporate such constraints in the proposed iLQR variant in following contents.
\subsection{Coarse Solution with Augmented Lagrangian Method}
In the first stage, we wrap up M-iLQR with Augmented Lagrangian (ALM-iLQR) as in \cite{HowellIROS2019}, then update the multipliers after one iteration of the inner M-iLQR solver. ALM-iLQR is explained in Algorithm \ref{alg:almddp}. 
\subsubsection{ALM-iLQR reformulation}
Based on (\ref{3}), we can write the augmented Lagrangian as below, 
\begin{subequations} 
    \begin{gather*}
    J_1(\boldsymbol{X}_{node},\boldsymbol{U};\boldsymbol{\lambda}, \mu) = J(\boldsymbol{X}_{node},\boldsymbol{U})\\
     + \sum_{k=0}^{N-1} (\boldsymbol{\lambda}_k^T \boldsymbol{h}_k + \frac{\mu_k}{2}||\boldsymbol{h}_k||^2) 
     + (\boldsymbol{\lambda}^T_N \boldsymbol{h}_N + \frac{\mu_N}{2}||\boldsymbol{h}_N||^2),
    \end{gather*}
\end{subequations}
where $\boldsymbol{h}=\max(\boldsymbol{0},\;\boldsymbol{g})$, $\boldsymbol{\lambda}$ is the dual variable, and $\mu$ is the penalty weight. The backward pass absorbs the derivatives of the penalty terms by fixing the multipliers. The recursive updates become,
\begin{subequations}
\begin{align}
     Q_{\boldsymbol{x},k}  & = \ell_{\boldsymbol{x},k} +\boldsymbol{f}_{\boldsymbol{x},k}^T V^{+}_{\boldsymbol{x},k+1}+{\boldsymbol{h}^T_{\boldsymbol{x},k}(\boldsymbol{\lambda}_k+{\mu_k}\boldsymbol{h}_k)},\\
     Q_{\boldsymbol{u},k}  & = \ell_{\boldsymbol{u},k} +\boldsymbol{f}_{\boldsymbol{u},k}^T V^{+}_{\boldsymbol{x},k+1}+{\boldsymbol{h}^T_{\boldsymbol{u}_k}(\boldsymbol{\lambda}_k+{\mu_k}\boldsymbol{h}_k)}, \\
     Q_{\boldsymbol{xx},k} & = \ell_{\boldsymbol{xx},k}+\boldsymbol{f}_{\boldsymbol{x},k}^T V_{\boldsymbol{xx},k+1}\boldsymbol{f}_{\boldsymbol{x},k}+{\mu_k}{\boldsymbol{h}^T_{\boldsymbol{x},k}\boldsymbol{h}_{\boldsymbol{x},k}}, \\
     Q_{\boldsymbol{uu},k} & = \ell_{\boldsymbol{uu},k}+\boldsymbol{f}_{\boldsymbol{u},k}^T V_{\boldsymbol{xx},k+1}\boldsymbol{f}_{\boldsymbol{u},k}+{\mu_k}{\boldsymbol{h}^T_{\boldsymbol{u},k}\boldsymbol{h}_{\boldsymbol{u},k}}, \\
     Q_{\boldsymbol{ux},k} & = \ell_{\boldsymbol{ux},k}+\boldsymbol{f}_{\boldsymbol{u},k}^T V_{\boldsymbol{xx},k+1}\boldsymbol{f}_{\boldsymbol{x},k}+{\mu_k}{\boldsymbol{h}^T_{\boldsymbol{u},k}\boldsymbol{h}_{\boldsymbol{x},k}}, \\
     Q_{\boldsymbol{xu},k} & = \ell_{\boldsymbol{xu},k}+\boldsymbol{f}_{\boldsymbol{x},k}^T V_{\boldsymbol{xx},k+1}\boldsymbol{f}_{\boldsymbol{u},k}+{\mu_k}{\boldsymbol{h}^T_{\boldsymbol{x},k}\boldsymbol{h}_{\boldsymbol{u},k}}.
\end{align}
\end{subequations}
All the constraints are linearized, which is the same as the linearization of dynamics.
\subsubsection{Update Augmented Lagrangian}
When one inner iteration is done with the fixed $\boldsymbol{\lambda}$ and $\mu$, the Lagrangian multipliers are updated as,
\begin{equation}\label{alupdate1}
\begin{aligned}
    \boldsymbol{\lambda}_{k} \gets \max(\textbf{0}, \boldsymbol{\lambda}_{k} + \mu_{k}\boldsymbol{g}_{k}(\boldsymbol{x}_k,\boldsymbol{u}_k)).
\end{aligned}
\end{equation}
and the penalty is increased by multiplying a scaling gain $\phi$.
\begin{equation}\label{alupdate2}
\mu_{k} \gets \phi \mu_{k}, \;  \phi>1.
\end{equation}
Augmented Lagrangian method is fast at first several iterations, so coarse trajectories can be obtained rapidly.
\subsection{Refined Solution with Relax Log Barrier}
The coarse solution from the first stage may not satisfy the tolerance of constraints or a high cost value due to the slow tailed convergence of the AL method \cite{Betts1998}, but it is still a good initial guess to warm start the second stage. To refine the solution, we apply the relaxed log barrier function method\footnote{The feasibility in the forward pass cannot be guaranteed due to $\boldsymbol{K}_{fb}$, which causes the penalty term to be undefined. Consequently, we use a relaxed extension.}, RLBM-iLQR, to approximate the original problem. RLBM-iLQR is outlined in Algorithm \ref{alg:rlbmddp}.
\subsubsection{Relaxed Log Barrier Function}
The relaxed log barrier function is
\begin{align*}
     B(g(x, u);\psi,\delta) = \left \{
     \begin{aligned}
        & -\psi ln(-g), \;\;  \delta \leq -g \\
        & \psi \beta (-g;\; \delta), \;\;-g \leq  \delta
        \end{aligned},~\psi > 0.
     \right.
\end{align*}
For $\beta$, we utilize a twice-differentiable quadratic extension as in \cite{HauserCDC2006},
\begin{align*}
     \beta (x;\;\delta) = \frac{1}{2}\left((\frac{x-2\delta}{\delta})^2-1 \right)-ln(\delta),\; \delta > 0.
\end{align*}
Adding the barrier term $B_k(\boldsymbol{g})$ to the objective function yields,
\begin{subequations}
    \begin{gather*}
    J_2(\boldsymbol{X}_{node},\boldsymbol{U};\psi,\delta)
    = J(\boldsymbol{X}_{node},\boldsymbol{U}) \\
    + \sum_{k=0}^{N-1} \sum_{i=1}^{q_k}B_{k}(\boldsymbol{g}_i(\boldsymbol{x}_k,\boldsymbol{u}_k;\psi,\delta)) + \sum_{j=1}^{q_N}B_N(\boldsymbol{g}_j(\boldsymbol{x}_N);\psi,\delta).
\end{gather*}
\end{subequations}
\subsubsection{Update RLB parameters}
To get a better approximation of the original constrained problem, we adapt a new strategy to update the parameters of the RLB function, 
\begin{subequations}\label{RLBupdate}
    \begin{gather}
    \psi \gets \omega_1 \psi, \; \psi > 0, \; \omega_1<1, \\
    \delta \gets \max(\delta_{min},\omega_2 \delta), \; \delta_{min} > 0, \; \omega_2<1.
    \end{gather}
\end{subequations}
The optimal solution is approached by decreasing $\psi$ and $\delta$ gradually. In fact, as $\psi \rightarrow 0$, the RLB function becomes a good approximation of the indicator function of the subset defined by the constraints. Initialized by the solution from ALM-iLQR, RLBM-iLQR obtains a polished solution with a better constraints satisfaction and can even refine an infeasible trajectory into a dynamically feasible one. The whole HM-iLQR algorithm is shown in Algorithm \ref{alg:hmddp}.
\begin{algorithm}[t]
    \caption{\lstinline{ALM-iLQR}}
    \label{alg:almddp}
    \begin{algorithmic}[1]
        \REQUIRE \small{$\boldsymbol{X}^{init}_{node}$, $\boldsymbol{U}_{init}$},\;$c^{al}_{min}$\\
        (\small{$c^{al}_{min}$: constraint tolerance in AL stage})
        \STATE \textbf{Initialize}: $\boldsymbol{\lambda}$, $\mu$, $\phi$, $c^{al}_{min}$
        \WHILE{$max(\boldsymbol{||g||})>c^{al}_{min}$}
            \STATE $[\boldsymbol{X}^{al}, \boldsymbol{U}^{al},\boldsymbol{K}^{al}] \gets \arg\min J_1(\boldsymbol{X}_{node},\boldsymbol{U};\lambda,\mu)$ using \lstinline{M-iLQR}
            \STATE \textit{If update}: update $\boldsymbol{ \lambda}$,\;$\mu$ using (\ref{alupdate1}-\ref{alupdate2})
        \ENDWHILE
        \RETURN $\boldsymbol{X}^{al}$, $\boldsymbol{U}^{al}$ and $\boldsymbol{K}^{al}$
    \end{algorithmic}
\end{algorithm} 
\begin{algorithm}[t]
    \caption{\lstinline{RLBM-iLQR}}
    \label{alg:rlbmddp}
    \begin{algorithmic}[1]
        \REQUIRE \small{$\boldsymbol{X}^{al}$, $\boldsymbol{U}^{al}$,\; $\boldsymbol{K}^{al}$,\;$c^{rlb}_{min}$} \\
        (\small{$c^{rlb}_{min}$: constraint tolerance in RLB stage})
        \STATE Initialize: $\delta$, $\omega_1$, $\omega_2$, $\psi$, $c^{rlb}_{min}$
        \WHILE{$max(\boldsymbol{||g||})>c^{rlb}_{min}$}
            \STATE $[\boldsymbol{X}_{sol}, \boldsymbol{U}_{sol},\boldsymbol{K}_{sol}] \gets \arg\min J_2(\boldsymbol{X},\boldsymbol{U},\boldsymbol{K};\psi, \delta)$ using \lstinline{M-iLQR}
            \STATE \textit{If update}: update $\psi$, $\delta$ using (\ref{RLBupdate})
        \ENDWHILE
        \RETURN $\boldsymbol{X}_{sol}$, $\boldsymbol{U}_{sol}$ and $\boldsymbol{K}_{sol}$
    \end{algorithmic}
\end{algorithm}
\begin{algorithm}[ht]
    \caption{\lstinline{HM-iLQR}}
    \label{alg:hmddp}
    \begin{algorithmic}[1]
        \REQUIRE Same as Algorithm 1
        \STATE \textbf{Initialize}: Set required parameters
        \STATE $[\boldsymbol{X}^{al}, \boldsymbol{U}^{al},\boldsymbol{K}^{al}] \gets$ \lstinline{ALM-iLQR}$(\boldsymbol{X}^{init}_{node}, \boldsymbol{U}_{init})$
        \STATE $[\boldsymbol{X}_{sol}, \boldsymbol{U}_{sol},\boldsymbol{K}_{sol}] \gets$ \lstinline{RLBM-iLQR}$(\boldsymbol{X}^{al}, \boldsymbol{U}^{al},\boldsymbol{K}^{al})$ 
        \RETURN $\boldsymbol{X}_{sol}$, $\boldsymbol{U}_{sol}$ and $\boldsymbol{K}_{sol}$
    \end{algorithmic}
\end{algorithm} 
\subsection{Improved Globalization Strategies}
In iLQR-like methods, the inaccurate approximation may cause the Hessian matrix to be not positive definite. Therefore, globalization strategies are required to improve the numerical stability. In this paper, we proposed an improved globalization strategy named CRLS which exploits the coupled effects of line-searching and regularization.
\subsubsection{Line Search (LS)}
When the new trajectory is updated with a full step, the state can stride out of the feasible approximated region. We adapt a backtracking line search method with Armijo condition\cite{Nocedal1999} in HM-iLQR. The system is integrated with,
\begin{equation}\label{FPLS}
    \hat{\boldsymbol{u}}_k = \boldsymbol{u}_k + \alpha \boldsymbol{k}_{ff,k} + \boldsymbol{K}_{fb,k}(\hat{\boldsymbol{x}}_k - \boldsymbol{x}_k),
\end{equation}
where $\alpha$ is a scalar parameter. HM-iLQR line-searches over an expected cost reduction as used in \cite{TassaIROS2012}, but an additional term related to the defects is suggested as below,
{\small
\begin{equation*}
    \begin{aligned}
    \Delta V &= \alpha \sum ^{N-1}_{i=0}  \boldsymbol{k}^T_{ff,i}Q_{\boldsymbol{u},i} + \frac{\alpha ^2}{2} \sum ^{N-1}_{i=0} \boldsymbol{k}^T_{ff,i}Q_{ \boldsymbol{uu},i} \boldsymbol{k}_{ff,i} + w\sum_{j=1}^M\boldsymbol{d}^T_j \boldsymbol{d}_j\\
    &= \alpha \Delta V_1 + \frac{\alpha ^2}{2} \Delta V_2 + w\sum_{j=1}^M\boldsymbol{d}^T_j\boldsymbol{d}_j,
\end{aligned}
\end{equation*}}
where $w$ is the weight of the additional cost term. $\Delta V$ plays a role of merit function which trades off between minimizing cost and closing defects. The ratio between the actual and expected reduction is,
\begin{equation*}
    ratio = ({J_{[\cdot]}(\hat{\boldsymbol{X}},\hat{ \boldsymbol{U}})-J_{[\cdot]}(\boldsymbol{X},\boldsymbol{U})})/{\Delta V},
\end{equation*}
where $J_{[\cdot]}$ is $J_{1}$ or $J_{2}$. The iteration is accepted if $ratio$ lies in a certain interval. Otherwise, $\alpha$ is reduced with a linear law,
\begin{equation}
    \alpha \gets \gamma \alpha,\; \gamma<1
\end{equation}
\subsubsection{Regularization Schedule (RS)} Regularization can enhance the stability of iLQR algorithms. We add a diagonal term to the Hessian of
state value as \cite{TassaIROS2012}, ,
\begin{subequations}
\begin{gather}
    \Tilde{Q}_{\boldsymbol{uu}}  =  \ell_{\boldsymbol{uu},k}+\boldsymbol{f}_{\boldsymbol{u},k}^T (V_{\boldsymbol{xx},k+1} + \mu_V \boldsymbol{I}_n) \boldsymbol{f}_{\boldsymbol{u},k}, \\
    \Tilde{Q}_{\boldsymbol{ux}}  =  \ell_{\boldsymbol{ux},k}+\boldsymbol{f}_{\boldsymbol{u},k}^T (V_{\boldsymbol{xx},k+1} + \mu_V \boldsymbol{I}_n) \boldsymbol{f}_{\boldsymbol{x},k}, \\
    \boldsymbol{k}_{ff}  = - \Tilde{Q}^{-1}_{\boldsymbol{uu}} Q_{\boldsymbol{u}}, \;
    \boldsymbol{K}_{fb}  = - \Tilde{Q}^{-1}_{\boldsymbol{uu}} \Tilde{Q}_{\boldsymbol{ux}}.
\end{gather}
\end{subequations}
The introduced term prevents the exponential growth of the negative eigenvalues of $V_{\boldsymbol{xx}}$ backwards in time and makes the update of state sequence more conservative.
\begin{algorithm}[bt]
    \caption{\lstinline{CRLS} Pseudocode}
    \label{alg:crls}
    \begin{algorithmic}[1]
        \REQUIRE $\sigma$, $\gamma$, $\mu^{max}_{V}$, $\mu^{0}_{V}$, $\alpha_{0}$, $\alpha_{min}$
        \STATE $flag^{\rm 1}\gets0,\;success^{\rm 2}\gets0$
        \STATE $\alpha \gets \alpha_0$, $\mu_V \gets \mu^0_V$
        \WHILE{$flag==0\; ||\; success==0$}
            \STATE $[\boldsymbol{k}_{ff},\boldsymbol{K}_{fb},\Delta V, flag] \gets$ \lstinline{Backward_Pass} with $\mu_V$
            \STATE go to line $12$ \textbf{if} $flag==0$
            \WHILE{$\alpha > \alpha_{min}$}
                \STATE $[\hat{\boldsymbol{X}},\hat{\boldsymbol{U}},\hat{J},success] \gets$ \lstinline{Forward_Pass} with $\alpha$
               \STATE break \textbf{if} $success == 1$
                \STATE $\alpha \gets \gamma \alpha$
            \ENDWHILE
            \IF{$success==0$ and $\mu_V < \mu^{max}_V$}
                \STATE  $\mu_V \gets \min(\sigma \mu_V, \mu^{max}_V)$, $\alpha \gets \alpha_0$
            \ELSE
                \STATE break
            \ENDIF
        \ENDWHILE
        \RETURN $\hat{\boldsymbol{X}}$, $\hat{\boldsymbol{U}}$, $\boldsymbol{k}_{ff}$ and $\boldsymbol{K}_{fb}$
    \end{algorithmic}
    \begin{tablenotes}
        \item $^{\rm 1}flag=1$ if backward pass succeeded.
        \item $^{\rm 2}success=1$ if forward pass succeeded.
    \end{tablenotes}
\end{algorithm} 
\subsubsection{Combination of Regularization and LS (CRLS)}
The regularization and LS techniques were used separately in previous works, like \cite{MartiIROS2020} \cite{TassaIROS2012}. They only added a regularization term if the backward pass failed (i.e., a non-positive definite $Q_{\boldsymbol{uu}}$ exits) or used LS for guaranteeing cost reduction. Failed LS would cause a termination of iterations in their algorithms.
However, $\alpha$ in LS controls the feedforward term, and a small value may cause $\boldsymbol{x}_N$ away from the goal state due to the shooting property of iLQR, especially when the trajectories are close to the local minimum. Hence, our solution is alternatively returning back to the backward pass and increasing the regularization $\mu_V$  when LS failed. A linear modification law for $\mu_V$ is proposed as below,
\begin{equation}
    \mu_V \gets \min(\sigma \mu_V, \mu^{max}_V).
\end{equation}
The pseudocode of CRLS is outlined in Algorithm \ref{alg:crls}.
A larger $\mu_V$ renders a higher $\boldsymbol{K}_{fb}$ to dominate and stabilize the next forward integration, which forces the state trajectory make a fine update. Such strategy can improve the convergence rate by avoiding small line-search factors. More details will be investigated in Section IV-B. Note that we only utilize CRLS in the RLB stage of HM-iLQR to ensure an adequate refinement of the coarse solution.
\section{Validation and Comparison}
We evaluated HM-iLQR on various constrained problems and conducted comparisons with different types of TO methods, including DIRCOL, direct single shooting (SS), AL-iLQR and ALTRO. First three problems use a quadratic objective function as below,
\begin{equation*}
    \begin{aligned}
        \ell(\boldsymbol{x}, \boldsymbol{u}) &= \frac{1}{2} 
        \left( 
        (\boldsymbol{x}-\boldsymbol{x}_{g})^T\boldsymbol{Q}(\boldsymbol{x}-\boldsymbol{x}_{g}) +
         \boldsymbol{u}^T\boldsymbol{R}\boldsymbol{u}
         \right) dt\\
        \ell(\boldsymbol{x}_f) &=\frac{1}{2} (\boldsymbol{x}_f-\boldsymbol{x}_{g})^T\boldsymbol{Q}_{f}(\boldsymbol{x}_f-\boldsymbol{x}_{g}) dt
    \end{aligned}
\end{equation*}
where $\boldsymbol{Q}$, $\boldsymbol{Q}_f$, and $\boldsymbol{R}$ are weight matrices. $dt$ is the time step. HM-iLQR was implemented with MATLAB and DIRCOL was implemented with an open-source trajectory optimization library of \textit{OptimTraj} \cite{Kelly2017}. \textit{OptimTraj} solved the NLP using \lstinline{SQP} \cite{Gill2005} in MATLAB. SS was implemented using the MATLAB interface of CasADi \cite{Andersson2019} with \lstinline{IPOPT} solver. ALTRO was mainly based on the open-source Julia code of \cite{HowellIROS2019}.
All computation was performed on a standard computer with an Intel(R)7-8700 CPU and 16GB RAM.

\subsection{Validation Examples}
\subsubsection{CartPole}
\begin{figure*}[htb]
    \centering
    \includegraphics[width=177mm]{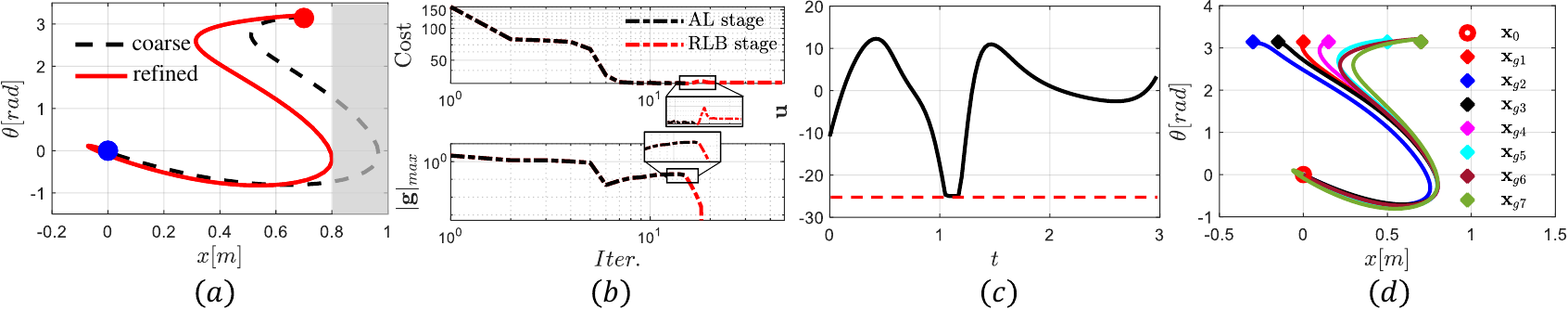}
    \caption{CartPole swinging up task. (a) $x-\theta$ trajectories. (b) Cost and constraint violation profiles. (c) Optimized control policy. (d)  Optimized $x-\theta$ trajectories for different goal positions.\\}
    \label{fig:cartpoleres}
\end{figure*}

The task is to swing up and move the CartPole \cite{Kelly2017} to different desired positions with limited force and rail range. Time horizon $T=3$ s and $N=100$. The shooting phase is $M=20$. The initial node states and control policy for this task are obtained from a dynamically infeasible solution from DIRCOL with a low discrete resolution ($N=10,\; Iter.=5$). We show the results of optimized state trajectory and control policy in Fig. \ref{fig:cartpoleres} (a) and Fig. \ref{fig:cartpoleres} (c). Fig. \ref{fig:cartpoleres} (b) shows the cost and constraint violations of the two stages during iterations. The first stage returned a coarse solution with a large constraint violation, even though the cost is low. In the RLB stage, the remained constraint violation was rapidly eliminated while there was a small cost increment as shown in Fig. \ref{fig:cartpoleres} (b). Fig. \ref{fig:cartpoleres} (d) shows the optimized trajectories for different desired horizon positions.
\subsubsection{2D Car}
\begin{figure*}[htb]
    \centering
    \includegraphics[width=176mm]{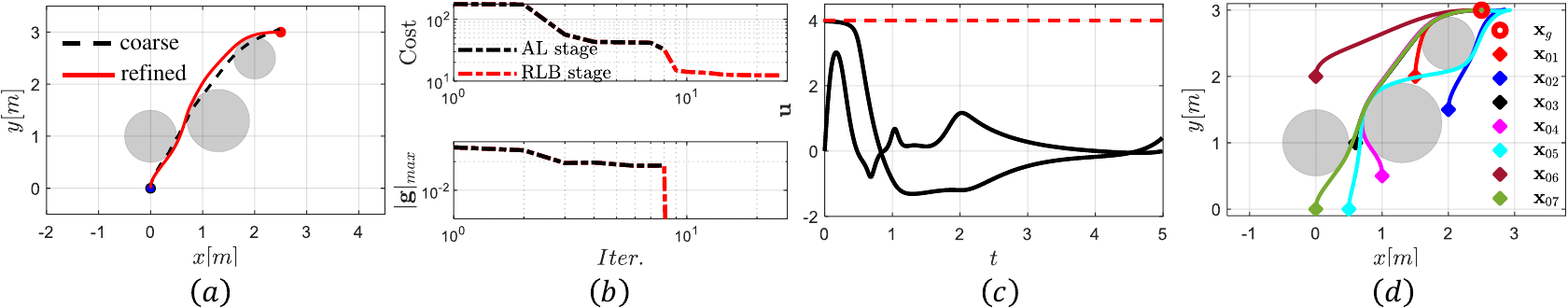}
    \caption{2D car motion planning task. (a) $x-y$ trajectories. (b) Cost and constraint violation profiles. (c) Optimized control inputs. (d) Optimized $x-y$ trajectories for different starting positions.\\}
    \label{fig:2dcareres}
\end{figure*}

In this example, a unit mass point with a discrete nonholonomic vehicle dynamics \cite{XieICRA2017} performs motion planning tasks of moving to the goal state $\boldsymbol{x}_g=[2.5,3.0,\pi/2,0,0,0]$ without hitting three circular obstacles. Limited actuation is also imposed. Time horizon $T=5$ s and $N=100$. The initialization for $\boldsymbol{X}^{init}_{node}$ is a linear interpolation ($M=10$) between $\boldsymbol{x}_0$ and $\boldsymbol{x}_g$. The initial control sequence is all zero. The simulation results are shown in Fig. \ref{fig:2dcareres}. In Fig. \ref{fig:2dcareres} (a), the car started from origin and moved to the goal position. As shown in Fig. \ref{fig:2dcareres} (b), the coarse trajectory (black dash line in Fig. \ref{fig:2dcareres} (a)) from AL stage were infeasible with a largest constraint violation of $0.07$. In the second stage, one more iteration was taken to eliminate the path constraints. The cost was also further reduced. For different starting positions, HM-iLQR obtained the optimal solutions stably as shown in Fig. \ref{fig:2dcareres} (d).

\subsubsection{2D Quadrotor}
\begin{figure*}[htb]
    \centering
    \includegraphics[width=176mm]{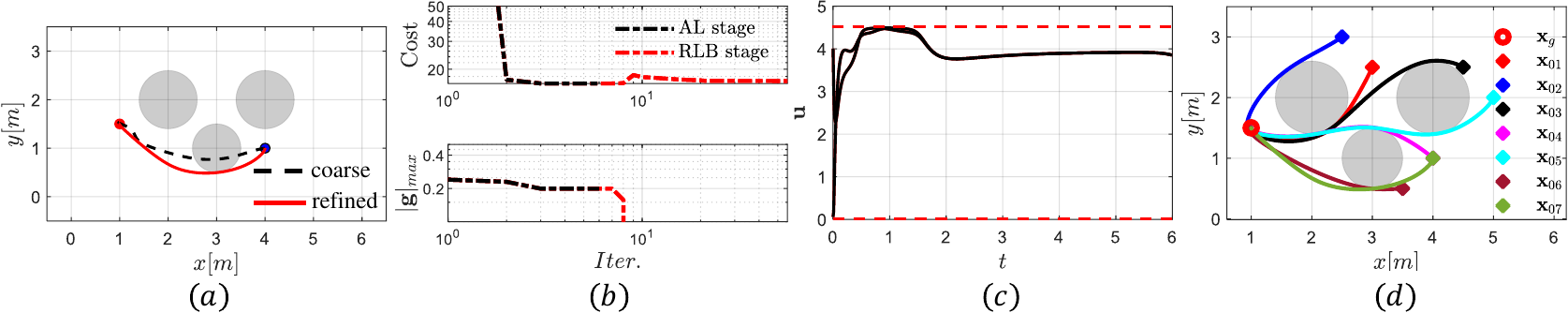}
    \caption{2D quadrotor motion plannning task. (a) $x-y$ trajectories. (b) Cost and constraint violation profiles. (c) Optimized control inputs. (d) Refined $x-y$ trajectories for different starting positions.}
    \label{fig:quadrotorres}
\end{figure*}

A planar quadrotor \cite{quadrotor} is an underactuated system with $\boldsymbol{x}=[x,y,\theta,\dot{x},\dot{y},\dot{\theta}]$ and $\boldsymbol{u}=[u_L,u_R]$. This task is to search for optimal collision-free trajectories reaching the goal position, $\boldsymbol{x}_g=[1.0,1.5,0,0,0,0]$, from different starting states with limited thrust forces. The path constraints include the limited body orientation (within $[-\frac{\pi}{6},\frac{\pi}{6}]$) and non-negative thrust forces. The time horizon $T=6$ s and $N=200$. The node states are initialized via a linear interpolation with $M=30$ and the initial policy is a hovering controller. 
The optimization details were shown in Fig. \ref{fig:quadrotorres}. 
We can see from Fig. \ref{fig:quadrotorres} (a) and Fig. \ref{fig:quadrotorres} (b) that the solution from AL stage was infeasible with a constraint violation up to $0.2$ although the objective value was low. By switching to the RLB stage, the constraint violations were eliminated within $3$ more iterations. We run several optimizations for other starting positions and the results were shown in Fig. \ref{fig:quadrotorres} (d).
\subsubsection{Robotic Manipulator}
\begin{figure}[bh]
    \centering
    \includegraphics[width=82mm]{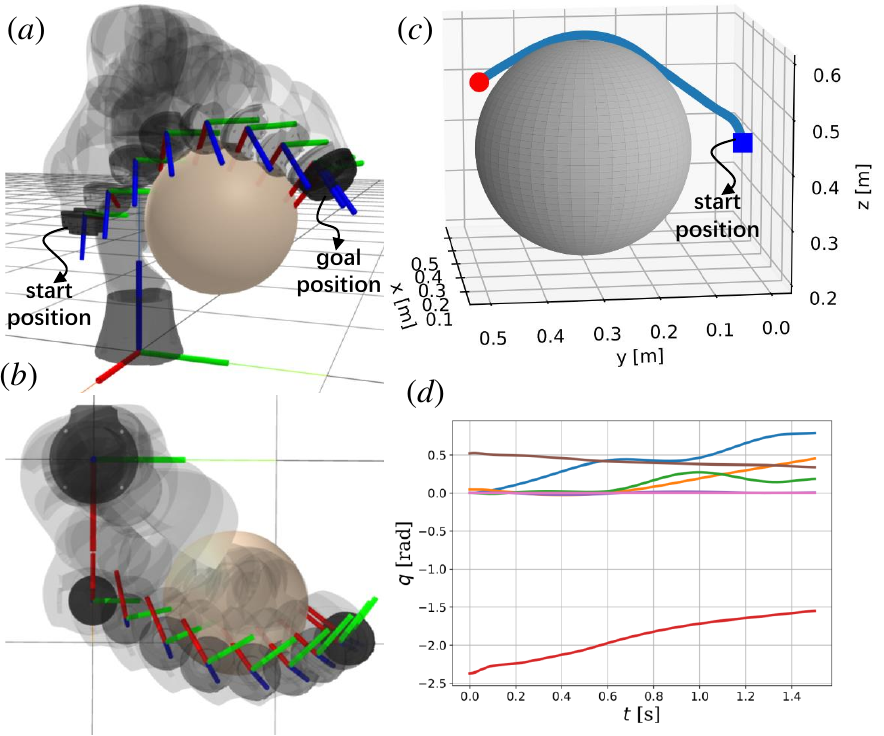}
    \caption{(a-b) Lateral view and top view of the optimized motion snapshots. (c) Collision-avoidance trajectory of the end-effector position. (d) Joint positions of the optimized motion.}
    \label{fig:iiwa}
\end{figure}

We validate HM-iLQR on a high-dimensional motion planning problem in this example, which is a constrained reaching task for a $7$ DoF industrial Kuka iiwa manipulator with considering the full kinematics and dynamics. The task is to move the end-effector to reach the target position, while avoiding the obstacle in the operational space (we only consider the collision between the obstacle and the last robot link here). For this manipulation task, the objective is,
\begin{equation*}
    \begin{aligned}
        \ell(\boldsymbol{x}, \boldsymbol{u}) &= \frac{1}{2} 
        \left( 
        \boldsymbol{e}^T\boldsymbol{H}\boldsymbol{e} + \boldsymbol{x}^T\boldsymbol{Q}\boldsymbol{x}+
         \boldsymbol{u}^T\boldsymbol{R}\boldsymbol{u}
         \right) dt\\
        \ell(\boldsymbol{x}_f) &=\frac{1}{2}
        \left(
        \boldsymbol{e}^T\boldsymbol{H}_{f}\boldsymbol{e} \right) dt
    \end{aligned}
\end{equation*}
where $\boldsymbol{x}=[\boldsymbol{q},\boldsymbol{\dot{q}}]^T$ and $\boldsymbol{H}$ is the weight matrix for the reaching task. $\boldsymbol{e}=\boldsymbol{p}_{ee}-\boldsymbol{p}_{g}$ represents the distance between the end-effector position $\boldsymbol{p}_{ee}$ and the target position $\boldsymbol{p}_{g}$, where $\boldsymbol{p}_{ee}$ is obtained via the forward kinematics, i.e. \lstinline{p_ee=fkin(q)}. The manipulator starts from $\boldsymbol{p}_{ee}=[0.3, 0.0, 0.4]$ and the target position is $\boldsymbol{p}_{g} = [0.35, 0.5, 0.5]$. The initial control guess is a gravity compensation controller. The time horizon\footnote{Here small time step, $dt$=$0.005$ s, helps to get stable forward simulations in high-dimensional nonlinear system.} $T=1.5$ s and $N=300$. The initialization for $\boldsymbol{X}^{init}_{node}$ is a linear interpolation with $M=5$ between the start and target configurations, which are obtained via inverse kinematics. The center of spherical obstacle is located at $\boldsymbol{p}_o=[0.3, 0.3, 0.4]$ with a radius $r_o=0.15$ m. For simplicity, the link of the end-effector is approximated by a ball with radius $r_{ee}=0.05$ m in the optimization. The collision-avoidance constraint is,
\begin{equation*}
    (r_o+r_{ee})^2-(\boldsymbol{p}_{ee} - \boldsymbol{p}_o)^T(\boldsymbol{p}_{ee} - \boldsymbol{p}_o) \leq 0.
\end{equation*}
Fig. \ref{fig:iiwa} (a-b) show the generated motion snapshots. The polished end-effector position of RLB stage in Fig. \ref{fig:iiwa} (c) shows the manipulator reached the target position without hitting the obstacle successfully.
\subsection{Local Convergence Analysis} 
\begin{figure}[bt]
    \centering
    \includegraphics[width=82mm]{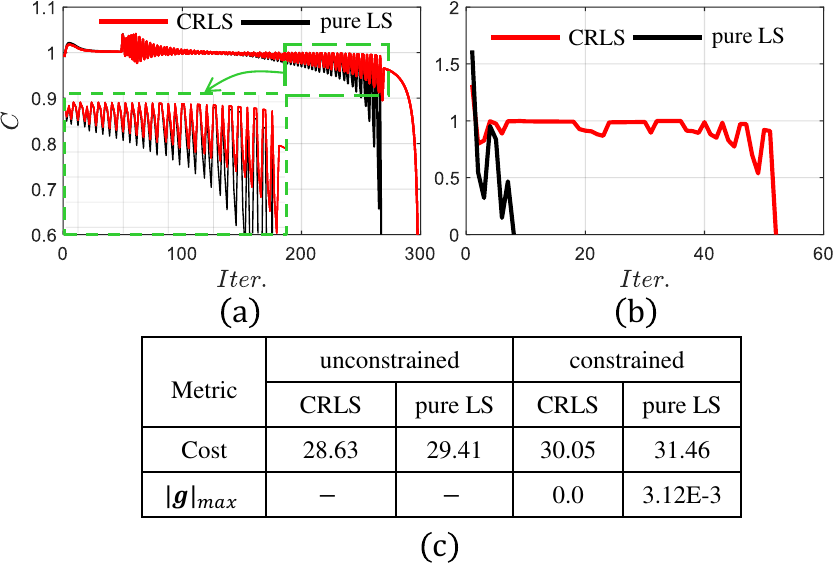}
    \caption{(a) Local convergence rates for unconstrained CartPole swinging up. (b) Local convergence rates for RLB stage of constrained CartPole swinging up. (c) Cost and constraint violation results for the four cases.}
    \label{fig:global_strat}
\end{figure}

We investigate the performance of the proposed combination of regularization and line search in terms of the local contraction rate of control policy with the CartPole example. The local convergence rate \cite{Nocedal1999} is,
\begin{equation}
    C=||^{k+1}\boldsymbol{U}-{^*}\boldsymbol{U}||_2/||^{k}\boldsymbol{U}-{^*}\boldsymbol{U}||_2,
\end{equation}
where $^{k}\boldsymbol{U}$ is the control policy at the $k$-th iteration and ${^*}\boldsymbol{U}$ is the converged optimal policy. We firstly illustrated CRLS with an unconstrained CartPole swinging up task ($M=20$) as shown in Fig. \ref{fig:global_strat} (a). The local contraction rates of CRLS and pure LS were similar at the first several iterations, because no LS or RS was required. However, when the solution was getting close to the optima, CRLS had a better local contraction property than pure LS as shown in the green dash box in Fig. \ref{fig:global_strat} (a). Pure LS terminated earlier and converged to a suboptimal solution with higher cost. This is in line with our discussion in Section III-C. Furthermore, CRLS obviously outperformed pure LS in the constrained case as shown in Fig. \ref{fig:global_strat} (b). LS got stuck with large constraint violations after a few iterations while CRLS fully eliminated the constraint violations as shown in Fig. \ref{fig:global_strat} (c). We can see that CRLS is more stable than LS, which is useful for the refinement of coarse solution. 

\subsection{Performance Comparison}
To further demonstrate the performance of HM-iLQR, we compared it against four TO methods in terms of cost, constraint feasibility and iterations. All problem settings were the same as Section IV-A.
The exit criteria is the same for all algorithms: threshold of objective function reduction: $\epsilon_v=1e{-3}$, constraint tolerance: $c^{rlb}_{min}=1e{-7}$, maximum iteration: $MaxIter=100$. The initialization strategies were the same as previous subsection. The initialization of ALTRO was the same as HM-iLQR. DIRCOL and SS were initialized with all-zero control policy. The initial state guess for DIRCOL was a linear interpolation between $\boldsymbol{x}_0$ and $\boldsymbol{x}_g$. We presented the comparisons for CartPole, 2D Car, and 2D quadrotor systems with two discrete resolutions in Tables \ref{table:cartpole}-\ref{table:2dquadrotor}.

\subsubsection{CartPole} Table \ref{table:cartpole} showed the details of different algorithms for the CartPole swinging up. HM-iLQR converged to fully feasible solutions with almost equally least costs for both discrete resolutions. By contrast, AL-iLQR failed with high objective values and large constraint violations. This was due to the numerical issues of AL approach that the large increased penalty terms cause severe numerical ill-conditioning. Although the costs of solutions from DIRCOL were close to HM-iLQR, it failed to satisfy the constraint tolerance within 100 iterations. Both SS and ALTRO reached the maximum iterations and got feasible solutions, but their costs were higher than HM-iLQR. Overall, HM-iLQR got the lowest costs and had a good performance in constraint satisfaction in this example. 

\subsubsection{2D Car}
We presented the comparisons of the 2D car example in Table \ref{table:2dcar}. AL-iLQR and SS showed similar performances as in CartPole system. Although ALTRO took the least iterations, the obtained solutions failed to satisfy the constraint tolerance. When $N=100$, HM-iLQR converged to a feasible suboptimal solution with the smallest cost value. DIRCOL obtained a similar solution as HM-iLQR but reached $MaxIter$. When $N=200$, algorithms except ALTRO reached the maximum iteration and HM-iLQR outperformed other approaches slightly in terms of cost. HM-iLQR performed better against ALTRO in handling constraints.

\subsubsection{2D Quadrotor} For planar quadrotor system, the initial state was set as $\boldsymbol{x}_0=[4.5, 2.5, 0.2,0,0,0]^T$ for benchmark comparisons. Table \ref{table:2dquadrotor} showed the comparison results. When $N=200$, HM-iLQR and SS obtained zero constraint-violated solutions. However SS converged to another suboptimal solution of a higher cost. ALTRO converged to a similar solution as HM-iLQR but with a larger objective value. AL-iLQR failed to find a feasible trajectory even though the cost was low. DIRCOL got the best solution which satisfied the constraint tolerance. With $N=300$, HM-iLQR obtained the best solution. AL-iLQR and SS converged to another suboptimal solution. While ALTRO stopped with a remained large constraint violation and high cost.

From the comparison results, we see that AL-iLQR was unstable and failed easily. HM-iLQR outperformed AL-iLQR in handling constraint and was more stable than approaches based on Augmented  Lagrangian method, i.e., AL-iLQR and ALTRO. ALTRO was fast but less stable in constraint satisfaction due to the active-set projections. Compared with SS, HM-iLQR had the advantage of initialization with infeasible trajectories. DIRCOL was a stable benchmark method and obtains competitive results in many optimization problems. However, DIRCOL was slower and took more iterations, which limits its application for online tasks. Last but not least, the retained feedback policy of HM-iLQR was also another advantage over other approaches for trajectory stabilization.

\section{Conclusion}
In this paper, we presented a Hybrid Multiple-shooting iterative Linear Quadratic Regulator approach to incorporate state and/or control inequality constraints. HM-iLQR could be accelerated with an infeasible initialization of the state trajectory due to the multiple-shooting framework. Furthermore, a two-stage framework was proposed by combining the features of AL approach and RLB method to achieve fast and high-precision constraint satisfaction. Then, a novel globalization strategy was introduced to improve the stability. We validated HM-iLQR and compared the performance with different approaches. The results showed that HM-iLQR was a stable and good constraint-satisfied nonlinear solver for constrained optimal control problems, and had good potentiality in the application of online predictive control. We did not compare the time performance of the algorithms as some of them were implemented using different programming languages. Future works include efficient implementations of HM-iLQR in C/C++ for high re-planning frequency in online tasks. Also, a sample-based planning planner could provide an initialization of intermediate states. Then HM-iLQR could be incorporated into a kinodynamic motion planner and be more applicable to high-dimensional systems.\\
\begin{table}[ht]
	\centering
	\caption{Benchmark comparisons of CartPole}
	\begin{tabular}{p{0.27cm}|p{0.7cm}|cccccccccccccccccccc}
    \toprule
    {$N$} &{Metric} &{{HM-iLQR}} &{AL-iLQR} &{DIRCOL} &{SS} &{ALTRO} \\
    \midrule
    \multirow{3}{2em}{{$50$}} 
     &Iter.  &$\boldsymbol{67}$  &$73$ &$100$     &$100$ &$100$ \\
     &Cost   &$\boldsymbol{29.46}$  &$49.09$ &$29.58$ &$34.62$ &$32.55$ \\
     &$|\boldsymbol{g}|_{max}$  &$\boldsymbol{0.0}$ &$0.06$ &$1.61$E-$4$ &$\boldsymbol{0.0}$ &$\boldsymbol{0.0}$\\
     \midrule
     \multirow{3}{2em}{$100$}
     &Iter. &$100$   &$\boldsymbol{61}$ &$100$ &$100$ &$100$  \\
     &Cost &$\boldsymbol{30.05}$   &$53.12$ &$30.12$ &$34.41$ &$32.62$\\
     &$|\boldsymbol{g}|_{max}$ &$\boldsymbol{0.0}$   &$\boldsymbol{0.0}$ &$1.73$E-$4$ &$\boldsymbol{0.0}$ &$\boldsymbol{0.0}$\\
    \bottomrule
    \end{tabular}
    \label{table:cartpole}
\end{table}

\begin{table}[ht]
	\centering
	\caption{Benchmark comparisons of 2D car}
	\begin{tabular}{p{0.27cm}|p{0.7cm}|ccccccccccccccccccc}
    \toprule
    $N$ &Metric &{HM-iLQR} &{AL-iLQR} &{DIRCOL} &{SS} &{ALTRO} \\
    \midrule
    \multirow{3}{2em}{$100$} 
     &Iter.  &$77$  &$84$ &$100$ &$100$ &$\boldsymbol{34}$    \\
     &Cost   &$\boldsymbol{12.03}$  &$13.13$ &$12.22$ &$13.01$ &$12.24$  \\
     &$|\boldsymbol{g}|_{max}$  &$\boldsymbol{0.0}$ &$2.61$E-$3$ &$9.58$E-$10$ &$\boldsymbol{0.0}$ &$2.91$E-$4$\\
     \midrule
     \multirow{3}{2em}{$200$} 
     &Iter. &$100$   &$100$ &$100$ &$100$ &$\boldsymbol{26}$    \\
     &Cost &$\boldsymbol{12.11}$   &$12.65$ &$12.21$ &$14.31$ &$12.16$  \\
     &$|\boldsymbol{g}|_{max}$ &$\boldsymbol{0.0}$   &$8.0$E-$3$ &$9.75$E-$10$ &$\boldsymbol{0.0}$ &$1.95$E-$4$\\
    \bottomrule
    \end{tabular}
    \label{table:2dcar}
\end{table}

\begin{table}[ht]
	\centering
	\caption{Benchmark comparisons of 2D quadrotor}
	\begin{tabular}{p{0.27cm}|p{0.7cm}|ccccc}
    \toprule
    $N$ &Metric &{HM-iLQR} &{AL-iLQR} &{DIRCOL} &{SS} &{ALTRO} \\
    \midrule
    \multirow{3}{2em}{$200$} 
     &Iter.  &$\boldsymbol{23}$  &$36$ &$100$ &$41$ &$28$    \\
     &Cost  &$22.31$ &$22.24$ &$\boldsymbol{22.09}$ &$23.47$ &$22.69$  \\
     &$|\boldsymbol{g}|_{max}$  &$\boldsymbol{0.0}$ &$3.68$E-$1$ &$1.76$E-$12$ &$\boldsymbol{0.0}$ &$6.25$E-$9$\\
     \midrule
     \multirow{3}{2em}{$300$} 
     &Iter. &$36$  &$28$ &$100$ &$53$ &$\boldsymbol{27}$   \\
     &Cost &$\boldsymbol{21.87}$   &$23.62$ &$21.93$ &$23.48$ &$23.80$  \\
     &$|\boldsymbol{g}|_{max}$   &$\boldsymbol{0.0}$ &$1.27$E-$9$ &$4.08$E-$5$ & $\boldsymbol{0.0}$ &$3.01$E-$4$\\
    \bottomrule
    \end{tabular}
    \label{table:2dquadrotor}
\end{table}


\begin{thebibliography}{}
\bibitem{Nocedal1999}
J. Nocedal and S. J. Wright, Numerical Optimization. {\em Springer}, 1999.

\bibitem{Betts1998}
Betts, John T. ``Survey of numerical methods for trajectory optimization." {\em Journal of guidance, control, and dynamics}, 21.2 (1998): 193-207.

\bibitem{Berts2014}
Bertsekas, Dimitri P. Constrained optimization and Lagrange multiplier methods. {\em Academic press}, 2014.

\bibitem{Stryk1992}
O. von Stryk and R. Bulirsch, ``Direct and indirect methods for trajectory optimization,” {\em Annals of Operations Research}, vol. 37, pp. 357–373, 1992.

\bibitem{Rao2009}
A. V. Rao, ``A survey of numerical methods for optimal control,” {\em Advances in the Astronautical Sciences}, vol. 135, no. 1, pp. 497–528, 2009.

\bibitem{Kelly2017}
Kelly, Matthew. ``An introduction to trajectory optimization: How to do your own direct collocation." {\em SIAM Review}, pp. 849-904, 2017.

\bibitem{Diehl2006}
Diehl, Moritz, et al. ``Fast direct multiple shooting algorithms for optimal robot control."  {\em Fast motions in biomechanics and robotics. Springer}, 65-93, 2006.

\bibitem{Mayne1966}
Mayne, David. ``A second-order gradient method for determining optimal trajectories of non-linear discrete-time systems." {\em International Journal of Control}, 3.1, 85-95, 1966.

\bibitem{iLQR2004}
Li, Weiwei, and Emanuel Todorov. ``Iterative linear quadratic regulator design for nonlinear biological movement systems." ICINCO (1). 2004.


\bibitem{GiftthalerIROS2018}
M. Giftthaler, M. Neunert, M. Stäuble, J. Buchli and M. Diehl, ``A Family of Iterative Gauss-Newton Shooting Methods for Nonlinear Optimal Control," {\em Proceedings of International Conference on Intelligent Robots and Systems}, pp. 1-9, 2018.


\bibitem{MastalliICRA2020}
C. Mastalli et al., ``Crocoddyl: An Efficient and Versatile Framework for Multi-Contact Optimal Control," {\em Proceedings of International Conference on Robotics and Automation}, pp.2536-2542, 2020 .

\bibitem{MartiIROS2020}
J. Marti-Saumell, J. Solà, C. Mastalli and A. Santamaria-Navarro, ``Squash-Box Feasibility Driven Differential Dynamic Programming," {\em Proceedings of International Conference on Intelligent Robots and Systems}, pp. 7637-7644, 2020.

\bibitem{TassaICRA2014}
Y. Tassa, N. Mansard and E. Todorov,``Control-limited differential dynamic programming," {\em Proceedings of International Conference on Robotics and Automation }, pp. 1168-1175, 2014.

\bibitem{TassaIROS2012}
Y. Tassa, T. Erez and E. Todorov, ``Synthesis and stabilization of complex behaviors through online trajectory optimization," {\em Proceedings International Conference on Intelligent Robots and Systems}, pp. 4906-4913, 2012.

\bibitem{XieICRA2017}
Z. Xie, C. K. Liu and K. Hauser, ``Differential dynamic programming with nonlinear constraints," {\em Proceedings of International Conference on Robotics and Automation}, pp. 695-702, 2017.

\bibitem{Yuichiro2020}
Aoyama, Yuichiro, et al. ``Constrained Differential Dynamic Programming Revisited." {\em arXiv preprint}, 2020.

\bibitem{Sarah2021}
Sarah Kazdadi, Justin Carpentier, Jean Ponce. ``Equality Constrained Differential Dynamic Programming.'' {\em Proceedings of International Conference on Robotics and Automation}, 2021.

\bibitem{LantoineAIAA2008}
Lantoine, Gregory, and Ryan Russell. ``A hybrid differential dynamic programming algorithm for robust low-thrust optimization." {\em AIAA/AAS Astrodynamics Specialist Conference and Exhibit}, 2008.

\bibitem{PavlovTCST2021}
Pavlov, Andrei, Iman Shames, and Chris Manzie. ``Interior point differential dynamic programming." {\em IEEE Transactions on Control Systems Technology}, 2021.


\bibitem{HowellIROS2019}
T. A. Howell, B. E. Jackson and Z. Manchester, ``ALTRO: A Fast Solver for Constrained Trajectory Optimization," {\em International Conference on Intelligent Robots and Systems}, pp.7674-7679, 2019.

\bibitem{Lantoine2012}
Lantoine, Gregory, and Ryan P. Russell. ``A hybrid differential dynamic programming algorithm for constrained optimal control problems. part 2: Application." {\em Journal of Optimization Theory and Applications}, 418-442, 2012.


\bibitem{HauserCDC2006}
J. Hauser and A. Saccon. A Barrier Function Method
for the Optimization of Trajectory Functionals with Constraints. {\em Proceedings of Conference Decision and
Control}, pp. 864–869, 2006.

\bibitem{quadrotor}
G. Wu and K. Sreenath, "Safety-critical control of a planar quadrotor," {\em Proceedings of American Control Conference}, pp. 2252-2258, 2016.

\bibitem{Neunert2016}
M. Neunert et al., "Fast nonlinear Model Predictive Control for unified trajectory optimization and tracking," {\em Proceedings International Conference on Robotics and Automation}, pp. 1398-1404, 2016.

\bibitem{Gill2005}
P. E. Gill, W. Murray, and M. A. Saunders, ``SNOPT: An SQP algorithm for large-scale constrained optimization," {\em SIAM review}, vol. 47, pp. 99–131, 2005.

\bibitem{Andersson2019}
Andersson, Joel AE, et al. ``CasADi: a software framework for nonlinear optimization and optimal control." {\em Mathematical Programming Computation}, 11.1, 1-36, 2019.
\end{thebibliography}
\end{document}